\documentclass[sn-mathphys-num]{sn-jnl}

\usepackage{graphicx}%
\usepackage{multirow}%
\usepackage{amsmath,amssymb,amsfonts}%
\usepackage{amsthm}%
\usepackage{mathrsfs}%
\usepackage[title]{appendix}%
\usepackage{xcolor}%
\usepackage{textcomp}%
\usepackage{manyfoot}%
\usepackage{booktabs}%
\usepackage{algorithm}%
\usepackage{algorithmicx}%
\usepackage{algpseudocode}%
\usepackage{listings}%
\usepackage{subfig}%

\begin{document}

\title[Article Title]{A Survey of Link Prediction in Temporal Networks}


\author*[1]{\fnm{Jiafeng} \sur{Xiong}}\email{jiafeng.xiong@manchester.ac.uk}
\author[2]{\fnm{Ahmad} \sur{Zareie}}\email{A.Zareie@sheffield.ac.uk}
\author[1]{\fnm{Rizos} \sur{Sakellariou}}\email{rizos@manchester.ac.uk}

\affil*[1]{\orgdiv{Department of Computer Science}, \orgname{University of Manchester}, \orgaddress{\street{Oxford Rd}, \city{Manchester} \postcode{M13 9PL}, \country{UK}}}
\affil[2]{\orgdiv{School of Computer Science}, \orgname{University of Sheffield}, \orgaddress{\street{211 Portobello}, \city{Sheffield} \postcode{S1 4DP}, \country{UK}}}


\abstract{Temporal networks have gained significant prominence in the past decade for modelling dynamic interactions within complex systems. A key challenge in this domain is Temporal Link Prediction (TLP), which aims to forecast future connections by analysing historical network structures across various applications including social network analysis. While existing surveys have addressed specific aspects of TLP, they typically lack a comprehensive framework that distinguishes between representation and inference methods. This survey bridges this gap by introducing a novel taxonomy that explicitly examines representation and inference from existing methods, providing a novel classification of approaches for TLP. We analyse how different representation techniques capture temporal and structural dynamics, examining their compatibility with various inference methods for both transductive and inductive prediction tasks. Our taxonomy not only clarifies the methodological landscape but also reveals promising unexplored combinations of existing techniques. This taxonomy provides a systematic foundation for emerging challenges in TLP, including model explainability and scalable architectures for complex temporal networks.}




\maketitle

\section{Introduction}\label{intro}

A temporal network is a network whose structure and properties evolve over time. In the past decade, temporal networks have been used to model interactions between the components of complex systems with dynamic structures and characteristics. A temporal network consists of two key elements: a set of nodes representing components of a system and a set of links indicating interactions between pairs of nodes. In temporal 
networks, the links between nodes may change over time; new nodes may also join the network. Identifying the likelihood of connecting between nodes to predict future links in a network (link prediction) is a fundamental problem that can be applied to various domains such as social networks~\cite{wang_link_2014}, telecommunication networks~\cite{borgnat_seven_2009}, traffic forecasting~\cite{vinchoff_traffic_2020,liu_foreseeing_2022} or knowledge graphs~\cite{nickel_review_2015,wang_survey_2021}.

Link prediction was first discussed in the context of static networks~\cite{lu_link_2011} where the aim is to predict future links based on a network's current state. In this paper this is referred to as traditional link prediction. However, the temporal link prediction (TLP) problem uses a network's historical states to predict the likelihood of future links. Incorporating historical dynamics into the TLP problem introduces additional complexity~\cite{holme_temporal_2023,kumar_link_2020}.

Recently, numerous methods have been proposed to address the TLP problem, each method focusing on different network characteristics. Solving TLP requires two key ingredients: first, a representation unit, which models how a temporal network can be captured to reflect its traits and historical dynamics; and second, an inference unit, which determines the likelihood of future links based on the learned representation. Since all TLP models necessarily need both units, this review is confined to approaches concerning either representation or inference. In the paper, TLP models are 
classified according to whether their primary contribution is to representation or inference and the defining attributes in each unit form the basis for the taxonomy adopted here. The survey begins by examining various representation methods and discusses their merits and challenges. It then classifies and evaluates the inference approaches that have been explored in the literature. Guided by this methodological perspective, the discussion is subsequently divided into two sections (Section~\ref{sec: literature1} and Section~\ref{sec: literature2}) dedicated to each unit respectively.

The field presents significant research opportunities through two primary avenues: (i) unexplored combinations of existing representation and inference units and (ii) the development of novel unit designs that could introduce new computational paradigms for either representation or inference. These opportunities could lead to more effective TLP approaches. 

Various surveys have been conducted to review TLP studies. Some surveys concentrate on specific network structures, such as homogeneous networks~\cite{qin_temporal_2022, xue_dynamic_2022},  or directed networks~\cite{ghorbanzadeh_hybrid_2021}. Other surveys focus on the application of TLP across various domains, encompassing temporal knowledge graphs~\cite{tian_knowledge_2022} and social networks~\cite{wang_link_2014, haghani_systemic_2019, daud_applications_2020}. Our survey sets itself apart in several aspects:

\begin{itemize} 
\item The survey introduces a novel taxonomy that distinguishes between the representation and inference units. Unlike existing surveys~\cite{qin_temporal_2022, xue_dynamic_2022, ghorbanzadeh_hybrid_2021} which classify models under a single hierarchical framework, our approach separates these components to provide deeper insights into the models’ applicability and performance variations whilst also exploring diverse network characteristics pertinent to the TLP problem.
\item The survey provides a comprehensive review of the network representation unit. While previous surveys~\cite{wang_link_2014, haghani_systemic_2019} provided some discussions of these models, we offer a systematic classification and analyse their ability to capture temporal and structural dynamics, thereby addressing a significant gap in the literature.
\item The survey identifies current challenges and outlines potential directions for future research in network representations, network characteristics, and TLP methodologies, offering valuable guidance for subsequent investigations in this field. 
\end{itemize}

Papers included in this survey were identified using Web of Science, Scopus, and Google Scholar and keywords on temporal networks, network representation and link prediction. Only  articles written in English in the period 2000–2024 that propose relevant algorithms or methods were considered. 

The rest of this paper is organised as follows. Section~\ref{sec: problem statements} presents the background on temporal graph representation and the problem statements. Following this, Section~\ref{sec: framework} introduces the taxonomy framework. Section~\ref{sec: literature1} and Section~\ref{sec: literature2} provide an overview of TLP methods, classifying them based on their representation and inference units. Section~\ref{sec: variations} and Section~\ref{sec: future} explore variations of the TLP problem and outline potential directions for future research. Finally, Section~\ref{sec: conclusion} presents the survey’s conclusion.


\section{Background}\label{sec: problem statements}
\subsection{Temporal Networks}

A temporal network represents interactions between components in a dynamic system over time. A dynamic system is composed of elements that interact with each other, and these interactions may evolve and change over time.

A temporal network can be modelled using a temporal graph $\mathcal{G = (V, E, T, X)}$~\cite{holme_temporal_2023}, where $\mathcal{V}$ is a set of nodes, and $\mathcal{E}$ is a set of edges denoting the interaction between pairs of nodes. The notation $\mathcal{T}$ and $\mathcal{X}$ indicate a time domain and a set of nodes' attributes, respectively. A temporal network can be defined as $\mathcal{G = (V, E, T)}$ if the nodes' attributes are ignored.

\subsection{Problem Statement}

\emph{Temporal Link Prediction (TLP)}: Given a temporal network $\mathcal{G = (V, E, T, X)}$ and a current timestamp $\tau \in \mathcal{T}$, TLP aims to predict future edges formed between nodes in the set $\mathcal{V}$ after timestamp $\tau$, based on the historical graph preceding $\tau$.

Unlike traditional link prediction, which focuses primarily on predicting links using node attributes and current topology, TLP requires models capable of capturing complex temporal and spatial network dynamics. This presents the challenging task of forecasting future link formation based on both historical and current network states.

\subsection{Representation Methods}
Temporal graphs can be described using two distinct approaches concerning the time dimension:
\begin{enumerate}
\item Discrete-time Dynamic Graphs (DTDG): A DTDG simplifies a temporal network into a sequence of timestamped snapshots, converting the continuous temporal space into discrete timestamps. This allows dynamic visualization to be transformed into a static, analysable format, though some information may be lost between snapshots during discretisation.
\item Continuous-time Dynamic Graphs (CTDG): This method abstains from discretizing the time dimension. Instead, it characterises the presence of all elements independently within a continuous temporal space, preserving the inherent temporal continuity of temporal networks.
\end{enumerate}

\subsubsection{Discrete-time Dynamic Graphs}
    \begin{figure} [htbp]
    \centering
        \includegraphics[width=.9\textwidth]{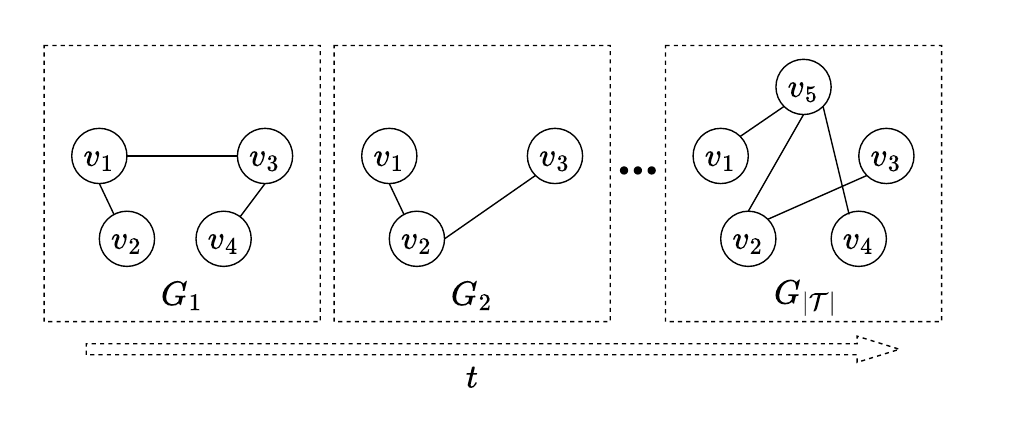}
        \caption{Discrete-time Dynamic Graphs (DTDG)} 
        \label{fig: Discrete Graph}
    \end{figure}

The discrete-time dynamic graphs define a temporal network $\mathcal{G = (V, E, T, X)}$ as a sequence of snapshots $\mathcal{G} = (G_1, G_2, \dots, G_{|\mathcal{T}|})$ over a discrete set of timestamps  $\mathcal{T}=\{1,2,\dots,{|\mathcal{T}|}\}$ where $|\cdot|$ is the size of the set.  Accordingly, the sequence of edges $\mathcal{E}$, nodes $\mathcal{V}$ and corresponding attributes $\mathcal{X}$ can be expressed as $\mathcal{E} = (E_1, E_2, \dots, E_{|\mathcal{T}|})$, $\mathcal{V} = (V_1, V_2, \dots, V_{|\mathcal{T}|})$ and  $\mathcal{X}=\{X_1, X_2, \dots, X_{|\mathcal{T}|}\}$. The interval between consecutive snapshots is presumed to be regular. 

Each snapshot at timestamp  $t$ can be represented as a tuple $G_t = (V_t, E_t, X_t)$, where $V_t = \{v^{t}_{1}, v^{t}_{2}, \dots, v^{t}_{N_t}\}$ denotes the set of nodes; $N_t$ is the number of nodes at timestamp $t$. $E_t = \{((v^t_{i},v^t_{j}), w_{ij})\mid v^t_{i},v^t_{j} \in V_t, w_{ij} \in \mathbb{R}_+\}$ denotes the set of edges and $((v^t_{i},v^t_{j}),w_{ij})$ represents the edge (also called link or event) between node $v^t_i$ and node $v^t_j$ and may be associated with a weight $w_{ij}$. An unweighted graph is a special case of a weighted graph with all edges weighting $1$. Similarly, the attribute set $X_t$ is $\{\phi(v^{t}_{1}),\phi(v^{t}_{2}),\dots,\phi(v^{t}_{N_t})\}$ and $\phi(\cdot)$ is the mapping function. Note that the attribute formation is usually a vector $\mathbf{x}_i^t \in\mathbb{R}^{d}$, and thus $X_t$ becomes an attribute matrix $\mathbf X_t \in \mathbb{R}^{N_t\times{d}}$. 

Moreover, the topology of snapshots can be expressed by an adjacency matrix set $\mathcal A = \{\mathbf{A}_1, \mathbf{A}_2, \dots, \mathbf{A}_t, \dots\}$. Here, $\mathbf{A}_t \in \mathbb{R}^{N_t \times N_t}$ corresponds to the snapshot $G_t$. For undirected unweighted graphs, $\mathbf{A}_t = \{a_{ij}^t\}_{i=1,j=1}^{N_t \times N_t}$ is a  symmetric binary matrix. Conversely, for undirected weighted graphs,  $\mathbf{A}_t$ is the same as the weight matrix whose elements are the weight of the edges.

Figure~\ref{fig: Discrete Graph} provides an example of the discrete-time dynamic graphs for a temporal graph. Each timestamp $t$ features a snapshot $G_t$ that depicts the system's behaviour, akin to user interactions in online social networks at a specific time. To minimise information loss, the interval between two adjacent snapshots should ideally equate to the minimum event duration. However, this can lead to significant data sparsity and redundancy in certain scenarios. For instance, if a changing email network is displayed as a series of snapshots, many snapshots might be empty or show very few connections~\cite{holme_modern_2015}. Consequently, this method often entails a trade-off between information loss and representational efficiency, which needs to be tailored to the specific research demands. The fundamental premise of this approach is to depict temporal networks as a sequence of static networks.

\subsubsection{Continuous-time Dynamic Graphs}
\begin{figure}[!ht]
  \centering
  \subfloat[Contact sequences]{\includegraphics[width=0.4\textwidth]{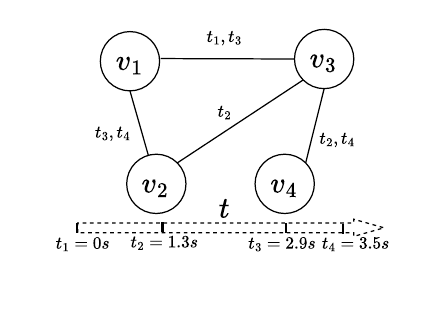}\label{fig: Contact sequence}}
  \hfill
  \subfloat[Interval graph]{\includegraphics[width=0.6\textwidth]{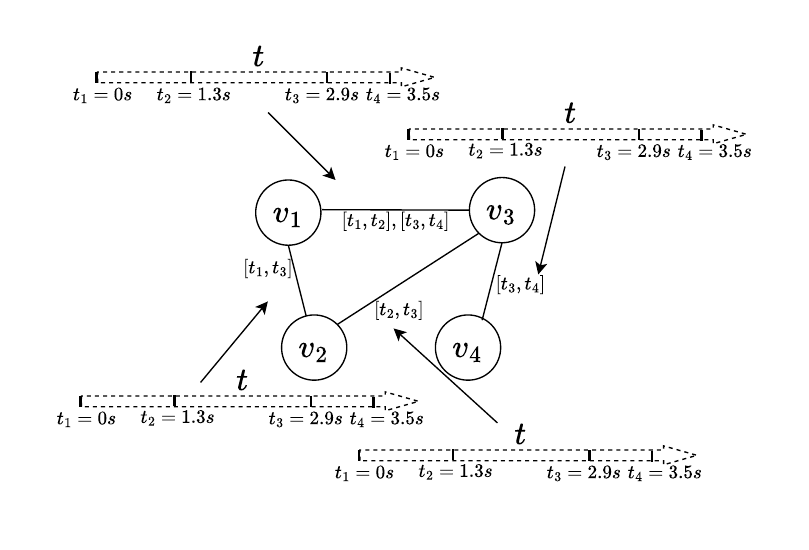}\label{fig: Interval graph}}
  \caption{Continuous-time Dynamic Graphs (CTDG)}
  \label{fig: Continuous Graph}
\end{figure}

Unlike discrete-time dynamic graphs, continuous-time dynamic graphs do not require equal time intervals between timestamps. For a period $T \subseteq \mathcal{T}$, a continuous-time dynamic graph $\mathcal{G = (V, E, T, X)}$ can be expressed as $G_T=(V_T,E_T,T, X_T)$, where
 $\mathcal{T}=\{t_1,t_2,\dots , t_{|\mathcal{T}|}\mid t_1<t_2<\dots<t_{|\mathcal{T}|}\}$ contains all timestamps throughout $\mathcal{G}$,
$V_T= \{v_1, v_2, \dots, v_{N_T}\}$ represents the nodes and
$X_T=\{\phi(v,t)\mid v\in V_T,t\in T\}$ denotes node attributes, with $\phi(\cdot)$ mapping nodes to their attributes.

Continuous-time dynamic graphs employ two principal methodologies to capture temporal network dynamics:
\begin{enumerate}
    \item \textbf{Contact sequences}: These record interactions as discrete events at specific timestamps, treating each interaction's duration as negligible. For instance, in an email network, edges represent instantaneous communication between nodes. This approach is also termed a graph stream~\cite{holme_modern_2015,divakaran_temporal_2020,10.1145/2627692.2627694,yu_netwalk_2018,ma_streaming_2020}. Formally, the edge set is defined as $E_T= \{((v_i,v_j), t, w_{ij}) \mid v_i, v_j \in V_T, w_{ij} \in \mathbb{R}_+, t \in T\} \subseteq \binom{V_T}{2} \times T \times \mathbb{R}_+$, indicating an edge $(v_i,v_j)$ with weight $w_{ij}$ exists at timestamp $t$. In Figure~\ref{fig: Contact sequence}, timestamps $t_2$ and $t_4$ mark when the edge $(v_3,v_4)$ was observed, with unequal intervals $[t_1,t_2]$ and $[t_2,t_3]$.
    
    \item \textbf{Interval graphs}: These account for the duration of interactions~\cite{divakaran_temporal_2020}, with time intervals explicitly representing how long connections persist. The edge set is defined as $E_T = \{((v_i,v_j),[t_a, t_b], w) \mid v_i, v_j \in V_T, w \in \mathbb{R}_+, t_a, t_b \in T, t_a<t_b\} \subseteq \binom{V_T}{2} \times \binom{T}{2} \times \mathbb{R}_+$. For example, the edge $((v_3,v_4),[t_3, t_4])$ in Figure~\ref{fig: Interval graph} indicates that this connection persists from $t_3$ to $t_4$.
\end{enumerate}

Although continuous-time dynamic graphs present greater challenges than discrete-time dynamic graphs when applying traditional static graph methods, they avoid the critical trade-off between information preservation and computational efficiency inherent in discrete-time approaches.

\subsubsection{Notation}
This survey uses $M$ to represent any temporal variable in temporal networks. The notation $M^e_s$ serves as a shorthand for variable $M$ in a temporal network from timestamp $s$ to timestamp $e$. Despite differences between discrete-time and continuous-time dynamic graphs regarding the temporal variable $M$, this survey unifies them through timestamp indices.
\begin{enumerate}
\item Discrete-time Dynamic graphs: The sequences of snapshots of the variable $M$ from timestamp $s$ to timestamp $e$ are denoted as $M^e_s=(M_{s}, M_{s+1}, ..., M_e)$.
\item Continuous-time Dynamic Graphs: The continuous-time representation of the variable $M$ from timestamp $s$ to timestamp $e$ is denoted as $M^e_s$.
\end{enumerate}
In addition, the rest of this survey uses $\tau$ to denote the index of the current timestamp, $\Delta$ as the forthcoming timestamps and $\Delta^\prime$ as the preceding timestamps.

\subsection{Inference Tasks}
\subsubsection{Transductive Tasks} 
A transductive task (or transductive inference) involves learning and predicting among observed data~\cite{vapnik_overview_1999}. In transductive TLP, the model predicts edges between observed nodes. Transductive TLP models $f_\text{TT}$ use the historical graph $G_{\tau - \Delta^{\prime}}^{\tau}$ from $(\tau - \Delta^{\prime})$ to $\tau$ and node attributes $X_{\tau - \Delta^{\prime}}^{\tau+\Delta}$ (if available) from $(\tau - \Delta^{\prime})$ to $(\tau+\Delta)$ as input to predict the future graph from $\tau$ to $(\tau + \Delta)$. The inference process of the transductive task can be represented as:
\begin{equation}
\hat{G}_{\tau}^{\tau+\Delta}= f_\text{TT}(G_{\tau - \Delta^{\prime}}^{\tau},X_{\tau - \Delta^{\prime}}^{\tau+\Delta})
\label{eq: transductive task}
\end{equation}
where $(\Delta \geq 1)$, $(\tau \geq \Delta^{\prime} \geq 0)$ and $\hat{G}^{\tau+\Delta}_{\tau}$ is the predicted graph. As the node set $\mathcal{V}$ has been observed during training, their information is encompassed within $G_{\tau - \Delta^{\prime}}^{\tau}$. 

\subsubsection{Inductive Tasks}
An inductive task (or inductive inference) pertains to the reasoning from observed training data to generalise to the unseen data~\cite{mitchell_need_2002}. Inductive TLP methods $f_\text{IT}$ use the historical graph $G_{\tau - \Delta^{\prime}}^{\tau}$ from timestamp $(\tau - \Delta^{\prime})$ to $\tau$, the node attributes $X_{\tau - \Delta^{\prime}}^{\tau+\Delta}$ (if available) from timestamp $(\tau - \Delta^{\prime})$ to $(\tau+\Delta)$, and future nodes $V_{\tau}^{\tau+\Delta}$ from timestamp  $\tau$ to $(\tau+\Delta)$ as input to predict the future graph $\hat{G}_{\tau}^{\tau+\Delta}$, induced by nodes $V_{\tau}^{\tau+\Delta}$. This can be articulated as:
\begin{equation}
\hat{G}_{\tau}^{\tau+\Delta}=f_\text{IT}(G_{\tau - \Delta^{\prime}}^{\tau},X_{\tau - \Delta^{\prime}}^{\tau+\Delta},V_{\tau}^{\tau+\Delta})
\label{eq: induction task}
\end{equation}
where $(\Delta \geq 1)$ and $(\tau \geq \Delta^{\prime}> 0)$.  TLP can be categorised into one-step tasks $(\Delta = 1)$ and multi-step tasks $(\Delta > 1)$ based on discrete-time dynamic graphs. Most existing methods primarily address one-step tasks.

Note that all the methods presented in this survey that can deal with induction tasks apply to transductive tasks but the reverse is not necessarily true.

\section{Taxonomy} \label{sec: framework}
\subsection{The Framework of the Taxonomy}
\begin{figure}[!ht]
  \centering
  \subfloat[Framework]{\includegraphics[width=0.33\textwidth]{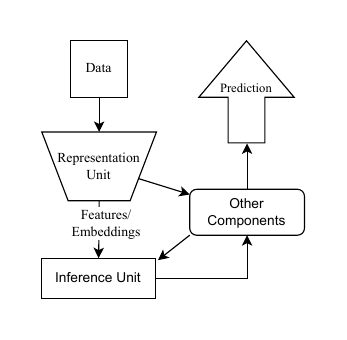}
  \label{fig: Framework}}
  \hfill
  \subfloat[Overview of the Taxonomy]{\includegraphics[width=0.65\textwidth]{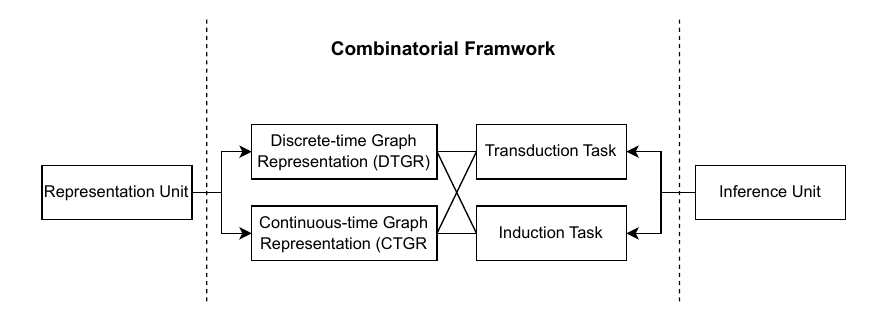}\label{fig: Taxonomy Main}}
  \caption{The Framework of the Taxonomy}
  \label{fig: Taxonomy}
\end{figure}

This survey proposes a taxonomy framework for TLP. It contends that all TLP methods comprise two sub-tasks: representing temporal networks and making predictions, which are processed by two distinct components: the Representation Unit (RU) and the Inference Unit (IU). The representation unit captures the temporal network structure and dynamics, whilst the inference unit leverages these representations to forecast future links. As depicted in Figure~\ref{fig: Framework}, nodes, edges, and attributes in temporal networks must undergo processing via a representation unit to feed them into the model and represent dynamic patterns. Subsequently, the inference units process features or embeddings derived from the representation unit, training a model to infer future links. In \cite{srinivasan_equivalence_2019}, transductive and inductive tasks relate to inference units but not to graph representation.

This survey categorises TLP methods based on their underlying elements, despite many approaches being hybrid, combining different representation and inference units or integrating general machine learning strategies. Figure~\ref{fig: Framework} shows each method must incorporate at least one representation unit and one inference unit. Since temporal networks can be modelled as either discrete-time or continuous-time dynamic graphs, and TLP tasks can be either transductive or inductive, methodologies are classified into four distinct categories based on these combinations, as illustrated in Figure~\ref{fig: Taxonomy Main}.
\begin{figure}[t]
\centering
    \includegraphics[width=1\textwidth]{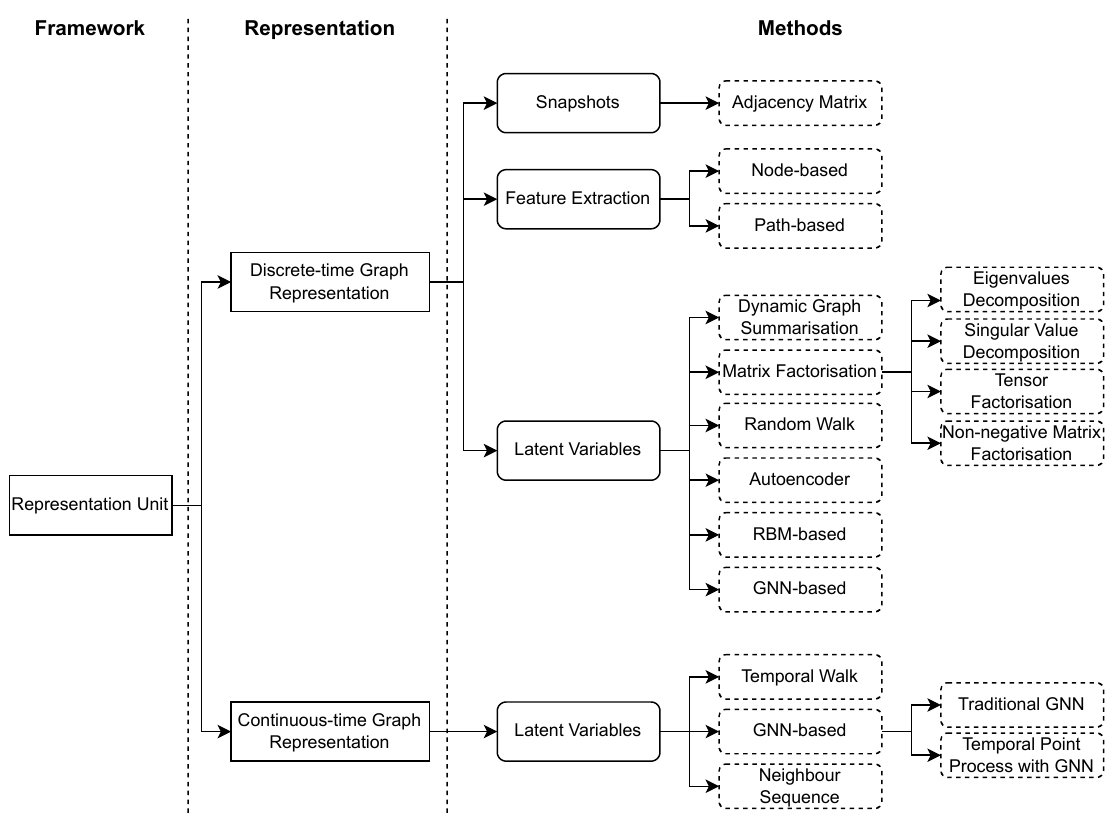}
    \caption{Taxonomy of Representation Unit}
    \label{fig: Representation Unit}
\end{figure}

\begin{table*}[!htbp]
\caption{Summary of Methods in Representation Unit. Key: GNN: Graph Neural Network; DGS: Dynamic Graph Summarisation; MF: Matrix Factorisation; RBM: Restricted Boltzmann Machine.}
\label{table: Representation Unit}
\begin{tabular*}{1\linewidth}{llll}
\toprule
Methods      & Representation & Inference & Representation Unit          \\
\midrule
Node-based          & Discrete             & Transductive   & Feature Extraction    \\
COMMLP-FULL~\cite{kumar_community-enhanced_2024}          & Discrete             & Transductive   & Feature Extraction    \\
CAES~\cite{choudhury_community-aware_2024}          & Discrete             & Transductive   & Feature Extraction    \\
Path-based          & Discrete             & Transductive   & Feature Extraction    \\
\hline
DGS~\cite{sharan_temporal-relational_2008}& Discrete             & Transductive   & DGS    \\
\hline
  ED~\cite{kunegis_network_2010}& Discrete             & Transductive   & MF                           \\
TSVD/CP~\cite{acar_link_2009}& Discrete             & Transductive   & MF                           \\
LIST~\cite{yu_link_2017}& Discrete             & Transductive   & MF                           \\
TMF~\cite{yu_temporally_2017}& Discrete             & Transductive   & MF                           \\
TBNS~\cite{yang_link_2012}& Discrete             & Transductive   & MF                           \\
DeepEye~\cite{ahmed_deepeye_2018}& Discrete             & Transductive   & MF                           \\
CRJMF~\cite{gao_temporal_2011}& Discrete             & Transductive   & MF                           \\
GrNMF~\cite{ma_graph_2018}& Discrete             & Transductive   & MF                           \\
SNMF-FC~\cite{ma_nonnegative_2017}& Discrete             & Transductive   & MF                           \\
AM-NMF~\cite{lei_adaptive_2018}& Discrete             & Transductive   & MF                           \\
\hline
DeepWalk~\cite{perozzi_deepwalk_2014}& Discrete             & Transductive   & Random Walk                  \\
Node2Vec~\cite{grover_node2vec_2016}& Discrete             & Transductive   & Random Walk                  \\
DynNode2Vec~\cite{mahdavi_dynnode2vec_2018}& Discrete             & Transductive   & Random Walk                  \\
SGNE~\cite{du_dynamic_2018}& Discrete             & Inductive      & Random Walk                  \\
\hline
SDNE~\cite{wang_structural_2016}& Discrete             & Transductive   & Autoencoder                  \\
Dyn-VGAE~\cite{mahdavi_dynamic_2020}& Discrete             & Transductive   & Autoencoder                  \\
DynGEM~\cite{goyal_dyngem_2018}& Discrete             & Inductive      & Autoencoder                  \\
\hline
ctRBM~\cite{li_deep_nodate}& Discrete             & Transductive   & RBM-based                     \\
\hline
CTDNE~\cite{nguyen_continuous-time_2018}& Continuous           & Transductive   & Temporal Walk          \\
\hline
Graph WaveNet~\cite{wu_graph_2019}& Discrete            & Transductive   & GNN-based                     \\
SDG~\cite{fu_sdg_2021}& Discrete             & Transductive   & GNN-based                    \\
TGN~\cite{rossi_temporal_2020}& Continuous           & Transductive   & GNN-based                     \\
TGGDN~\cite{huang_temporal_2023}& Continuous           & Transductive   &GNN-based                   \\
CoDyG~\cite{chen_correlation-enhanced_2024}& Continuous           & Transductive   &GNN-based                   \\
HTNE~\cite{zuo_embedding_2018}& Continuous           & Transductive   & GNN-based        \\
MTNE~\cite{huang_motif-preserving_2020}& Continuous           & Transductive   & GNN-based        \\
M\textsuperscript{2}DNE~\cite{lu_temporal_2019}& Continuous & Transductive   & GNN-based        \\
DyGCN~\cite{cui_dygcn_2022}& Discrete             & Inductive      & GNN-based                    \\
GCN-MA~\cite{mei_dynamic_2024} & Discrete             & Inductive      &GNN-based                \\
TGAT~\cite{xu2020inductiverepresentationlearningtemporal} & Continuous           & Inductive      & GNN-based  \\
TREND~\cite{wen_trend_2022}& Continuous           & Inductive      & GNN-based        \\
\hline
GraphMixer~\cite{cong_we_2023}& Continuous    & Inductive      & Neighbour Sequence        \\
DyGFormer~\cite{yu_towards_2023}& Continuous           & Inductive      & Neighbour Sequence      \\
FreeDyG~\cite{tian_freedyg_2024}& Continuous           & Inductive      & Neighbour Sequence      \\
\bottomrule
\end{tabular*}
\end{table*}

\subsection{Representation Unit}

Temporal networks can be represented as discrete-time or continuous-time dynamic graphs, each of which needs different graph representation capabilities to capture spatial and temporal information. Representation unit methodologies are used to represent real-world network data and extract features or latent information.
 
The representation unit fundamentally encapsulates the spatial dynamics within the temporal networks. As illustrated in Figure~\ref{fig: Representation Unit}, three principal methodologies are identified for the processing of graph dynamics in TLP:
\begin{enumerate}
    \item Snapshots: This approach necessitates directly feeding network snapshots into the predictive model. Because each snapshot can be considered as a static network and structurally depicted via an adjacency matrix $\mathcal{A}$, the Snapshots technique is commonly employed within discrete-time dynamic graphs.
    \item Features Extraction: This approach entails extracting explicit features from the node, edge, or entire graph levels such as similarity or centrality degree. These features are then used as input for the model. Features extraction is often employed to gather complex information beyond Snapshots to enhance the performance of models.
    \item Latent Variable: It represents latent variables that echo underlying factors and they are typically used to reduce the dimensionality and suppress noise within temporal networks. These approaches are commonly used when graph dynamics are highly complex and not easily summarised by explicit feats. Discrete-time and continuous-time dynamic graphs employ distinct approaches. However, all methods for continuous-time dynamic graphs fall under latent variables while discrete-time dynamic graphs methods encompass all three types mentioned above.
\end{enumerate}

Table~\ref{table: Representation Unit} catalogues the representation unit methods referenced in this survey, which are examined thoroughly in Section~\ref{sec: literature1}. To address TLP, researchers have proposed various representation units that fall into two categories: task-dependent units (specifically designed for TLP) and task-independent units (applicable across diverse downstream tasks in temporal networks). The process of generating latent variables through these representation units is broadly termed Dynamic Network Embedding (DNE)~\cite{barros_survey_2021,xue_dynamic_2022,qin_temporal_2022}.
DNE maps nodes to high-dimensional vector representations whilst preserving the dynamic patterns of topology and attributes. Formally, DNE learns a function $f_{\text{DNE}} : \mathcal{V} \mapsto \mathbb{R}^d, d < N$ that projects the node set $\mathcal{V}$ into a lower-dimensional space. 

In TLP applications, task-dependent DNE methods are optimised specifically for link prediction, yielding superior performance but limited transferability. In contrast, task-independent DNE methods create versatile embeddings suitable for multiple downstream tasks, including node classification and community detection, though they may deliver comparatively lower performance on TLP tasks.

\begin{figure} [t]
    \makebox[\textwidth][c]{\includegraphics[width=0.8\textwidth]{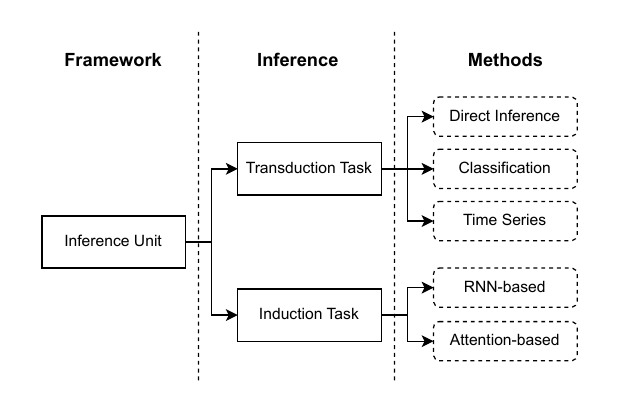}}
    \caption{Taxonomy of Inference Unit} 
    \label{fig: Inference Unit}
\end{figure}

\begin{table*}[!htbp]
\centering
\footnotesize
\caption{Summary of Methods in Inference Unit. Key: GNN: Graph Neural Network; DGS: Dynamic Graph Summarisation; MF: Matrix Factorisation; RBM: Restricted Boltzmann Machine; RNN: Recurrent Neural Network.}
\label{table: Inference Unit}
\begin{tabular*}{1\linewidth}{lllll}
\toprule
Methods       & Representation & Inference      & Representation Unit     & Inference Unit      \\
\midrule
TVRC~\cite{sharan_temporal-relational_2008}& Discrete       & Transductive  & DGS     & Classification      \\
GTRBM~\cite{li_restricted_2018}& Discrete       & Transductive  & RBM-based               & Classification      \\
\hline
TSalton~\cite{zhang_temporal_2020}& Discrete       & Transductive  & Node-based              & Time Series         \\
SR~\cite{fang_graph_2010}& Discrete       & Transductive  & MF                      & Time Series         \\
TRMF~\cite{yu_temporal_2016}& Discrete       & Transductive  & MF                      & Time Series         \\
ARIMA–Kalman ~\cite{xu_real-time_2017}& Continuous     & Transductive  & Snapshots               & Time Series         \\
\hline
DDNE~\cite{li_deep_2018}& Discrete       & Transductive  & Snapshots               & RNN-based           \\
DynGraph2Vec~\cite{goyal_dyngraph2vec_2020}& Discrete       & Transductive  & Autoencoder             & RNN-based           \\
EvolveGCN~\cite{pareja_evolvegcn_2020}& Discrete       & Transductive  & GNN-based               & RNN-based           \\
GCN-GAN~\cite{lei_gcn-gan_2019}& Discrete       & Transductive  & GNN-based               & RNN-based           \\
NetworkGAN~\cite{yang_advanced_2020}& Discrete       & Transductive  & GNN-based               & RNN-based           \\
DGNN~\cite{ma_streaming_2020}& Continuous     & Inductive     & GNN-based               & RNN-based           \\
CAW-N~\cite{wang_inductive_2022}& Continuous     & Inductive     & Temporal Random Walk    & RNN-based           \\
\hline
GAT~\cite{velickovic_graph_2018}& Discrete       & Inductive     & GNN-based               & Attention-based     \\
DySAT~\cite{sankar_dysat_2020}& Discrete       & Inductive     & GNN-based               & Attention-based     \\
ASTGCN~\cite{guo_attention_2019}& Discrete       & Inductive     & GNN-based               & Attention-based     \\
STGSN~\cite{min_stgsn_2021}& Discrete       & Inductive     & GNN-based               & Attention-based     \\
NeiDyHNE~\cite{wei_neighbor-enhanced_2024}& Discrete       & Inductive     & GNN-based               & Attention-based     \\
MAGNA~\cite{wang_multi-hop_2020}& Continuous     & Inductive     & GNN-based               & Attention-based     \\
DyRep~\cite{trivedi_dyrep_2019}& Continuous     & Inductive     & GNN-based  & Attention-based     \\
\bottomrule
\end{tabular*}
\end{table*}

\subsection{Inference Unit}

Modelling temporal dynamics is related to transductive and inductive tasks as shown in Figure~\ref{fig: Inference Unit}, which demands different abilities to integrate the upstream representation and learn temporal patterns for TLP. Inference units are the methods employed to process graph representations derived from representation units and learn dynamic patterns of temporal networks for TLP.

Direct inference is a prevalent approach in representation unit applications, whereby embeddings or features from dynamic network embedding methods directly yield predictions without requiring additional model training. Techniques such as Euclidean Distance or adjacent matrix transformations are commonly employed. The method's straightforward nature makes it particularly popular for one-step tasks.

Moreover, it is important to highlight that several inference methods, including RNN-based and attention-based models, initially designed for inductive tasks, can effectively handle transductive tasks. This shows the adaptability of these models and suggests the potential for their broader application in different TLP scenarios. Table~\ref{table: Inference Unit} summarises all the methods of inference units which are elaborated in Section~\ref{sec: literature2}

\section{Review of Representation Units for TLP}\label{sec: literature1}
\subsection{Discrete-time Dynamic Graphs Methods}
\subsubsection{Snapshots}

The Snapshots approach is a fundamental method employed in the domain of discrete-time dynamic graphs for TLP. This technique involves directly inputting network temporal adjacency matrices into the inference unit. The snapshots are treated as a sequence of static networks and are structurally represented by adjacency matrices. This method is effective for tasks that resemble time-series inference. The direct use of snapshots simplifies the representation and processing of dynamic graphs, making it an efficient choice for many application scenarios.

\subsubsection{Feature Extraction}
\paragraph{Node-based Similarity Approaches}

Entities tend to form new connections with those highly similar to themselves. Node neighbourhood is a crucial factor in similarity calculation. Building on this natural observation, researchers have developed numerous neighbour-based methods that utilise neighbourhood topological information from discrete snapshots for TLP. In this survey, $\Gamma(v_i^\tau)$ represents the set of neighbours of the node $v_i^\tau$, and $|\Gamma(v_i^\tau)|$ denotes the degree of the node $v_i^\tau$ (the number of its neighbours).

\noindent
\textbf{Common Neighbours (CN)}~\cite{Newman2001ClusteringAP} It measures the number of shared neighbours of two nodes $v_i^\tau$ and $v_j^\tau$. The higher the number of common neighbours, the higher the likelihood of a link between the nodes:
\begin{equation}
    \mathrm{CN}(v_i^\tau,v_j^\tau)\equiv|\Gamma(v_i^\tau)\cap \Gamma(v_j^\tau)|
    \label{eq: CN}
\end{equation}
\textbf{Preferential Attachment (PA)}~\cite{barabasi_evolution_2002}  This index suggests that the likelihood of a new connection between two nodes is proportional to the product of their degrees. It is often used in growing scale-free networks: 
\begin{equation}
    \mathrm{PA}(v_i^\tau,v_j^\tau)\equiv|\Gamma(v_i^\tau)|\cdot|\Gamma(v_j^\tau)|
    \label{eq: PA}
\end{equation}
\textbf{Jaccard Coefficient (JC)}~\cite{Goodall1978} The JC measures the common neighbours by the total number of unique neighbours of both nodes:
\begin{equation}
    \mathrm{JC}(v_i^\tau,v_j^\tau)\equiv\frac{|\Gamma(v_i^\tau)\cap \Gamma(v_j^\tau)|}{|\Gamma(v_i^\tau)\cup \Gamma(v_j^\tau)|}
    \label{eq: JC}
\end{equation}
\textbf{Sorensen Index (SI)}~\cite{Srensen1948AMO} The SI compares the number of common neighbours to the sum of the degrees of both nodes. It is more robust than the Jaccard Coefficient against outliers:
\begin{equation}
    \mathrm{SI}(v_i^\tau,v_j^\tau)\equiv\frac{2|\Gamma(v_i^\tau)\cap \Gamma(v_j^\tau)|}{|\Gamma(v_i^\tau)| + |\Gamma(v_j^\tau)|}
    \label{eq: SI}
\end{equation}
\textbf{Cosine Similarity (CS)}~\cite{10.1145/361219.361220} The CS measures the similarity between two nodes by calculating the cosine of the angle between their neighbour vectors: 
\begin{equation}
    \mathrm{CS}(v_i^\tau,v_j^\tau)\equiv\frac{|\Gamma(v_i^\tau)\cap \Gamma(v_j^\tau)|}{\sqrt{|\Gamma(v_i^\tau)|\cdot|\Gamma(v_j^\tau)|}}
    \label{eq: SC}
\end{equation}
\textbf{Adamic/Adar (AA)}~\cite{adamic_friends_2003}  AA is based on shared features; it assigns more weight to common neighbours with a smaller degree value: 
\begin{equation}
    \mathrm{AA}(v_i^\tau,v_j^\tau)\equiv\sum_{v_p^\tau \in \Gamma(v_i^\tau)\cap \Gamma(v_j^\tau)}\frac{1}{\log |\Gamma(v_p^\tau)|}
    \label{eq: AA}
\end{equation}

Node-based similarity approaches typically serve as the foundation for many community-based TLP models, such as COMMLP-FULL~\cite{kumar_community-enhanced_2024} and CAES~\cite{choudhury_community-aware_2024}. These methods offer significant advantages in terms of interpretability.

\paragraph{Path-based Similarity Approaches}

Path-based methods provide another perspective for measuring the similarity between nodes in a network. These methods concentrate on the length and quantity of paths between nodes. 
Unlike node-based approaches, path-based methods view networks as a series of connected paths, enabling the identification of similarities beyond immediate neighbourhoods. Some path-based methods also incorporate random Walks to account for the uncertainty and temporal evolution of networks.

In the rest of the survey,  $(\mathbf{M})_{a,b}$  or $(\mathcal{M})_{a,b,c}$ are denoted as elements in the matrix $\mathbf{M}$ or tensor $\mathcal{M}$ in the $a$-th row, $b$-th column, and $c$-th depth, respectively. $\mathbf m_{a}$ represents the $a$-th vector of the matrix $\mathbf{M}$. 

\textbf{Shortest Path (SP)}~\cite{10.1145/956863.956972} The SP converts the shortest path length between nodes into a similarity measure by taking its negative value. This transformation ensures that shorter paths correspond to higher similarity scores, while longer paths indicate lower similarity.
\begin{equation}
    \mathrm{SP}(v_i^\tau,v_j^\tau)\equiv-|d(v_i^\tau, v_j^\tau)|
\label{eq: SP}
\end{equation}
where $d(\cdot)$ is the shortest path between the node pair $(v_i^\tau, v_j^\tau)$ computed by the Dijkstra algorithm~\cite{dijkstra_note_1959}.

\textbf{Local Path (LP)}~\cite{PhysRevE.80.046122} LP makes use of information on local paths between node $v_i^\tau$ and node $v_j^\tau$ with a $2, 3$ and $4$-length rather than the nearest paths. LP suggests that $2$-length paths should have greater significance than $3$-length paths and $3$-length paths in relation to $4$-length paths. To account for this, an adjustment factor $\alpha$ is applied.  
\begin{equation}
    \mathrm{LP}(v_i^\tau,v_j^\tau)\equiv\mathbf{A}_\tau^{2}+\alpha \mathbf{A}_\tau^{3}+\alpha^{2}\mathbf{A}_\tau^{4},
    \label{eq: LP}
\end{equation}
where $\mathbf{A}_\tau^{2}$, $\mathbf{A}_\tau^{3}$ and $\mathbf{A}_\tau^{4}$ denote adjacency matrices about the nodes having $2, 3$ and $4$-length distances at timestamp $\tau$, respectively. The parameter $\alpha$ is close to $0$.

\textbf{Katz Index (KI)}~\cite{katz_new_1953} The KI measures similarity based on the number of paths of different lengths between two nodes. Shorter paths have larger similarities.
\begin{align}
    \mathrm{KI}(v_i^\tau,v_j^\tau) &\equiv \sum_{\ell=1}^\infty \beta^\ell |{paths}^{\ell}(v_i^\tau,v_j^\tau)| \notag 
    = \left( \sum_{\ell=1}^\infty \beta^\ell \mathbf{A}_\tau^\ell \right)_{i,j} \notag 
    = \left( \mathbf{I} - \beta \mathbf{A}_\tau \right)^{-1}_{i,j} - \mathbf{I}_{i,j}
    \label{eq: KI}
\end{align}
where ${paths}^{\ell}(v_i^\tau,v_j^\tau)$ is the set of total $\ell$-length paths between nodes $v_i^\tau$ and $v_j^\tau$. $\beta$ is a damping factor, which constrains the path weights and $\mathbf I$ is the identity matrix.

\textbf{Cosine Similarity Time (CST)}~\cite{fouss_random-walk_2007}
The CST is a variation of Cosine Similarity:
\begin{equation}
    \mathrm{CST}(v_i^\tau,v_j^\tau)\equiv\frac{(\mathbf L_{\tau}^{\dagger})_{i,j}}{\sqrt{(\mathbf L_{\tau}^{\dagger})_{i,i}(\mathbf L_{\tau}^{\dagger})_{j,j}}}
    \label{eq: CST}
\end{equation}

\textbf{SimRank (SR)}~\cite{10.1145/775047.775126}
The SR supposes that two nodes are considered similar if they are connected to similar nodes. This method calculates similarity based on the probability that two random walkers, starting from nodes $v_i^\tau$ and $v_j^\tau$, will meet at the same node in the future. The walkers move to a random neighbour at each step and the similarity is computed recursively.
\begin{equation}
    \mathrm{SR}(v_i^\tau,v_j^\tau)\equiv\left\{\begin{array}{lr}1&v_i^\tau=v_j^\tau\\ \alpha \cdot\frac{\sum_{v_p^\tau \in T(v_i^\tau)}\sum_{v_q^\tau \in T(v_j^\tau)} \mathrm{SR}(v_p^\tau,v_q^\tau)}{|\Gamma(v_i^\tau)|\cdot|\Gamma(v_j^\tau)|}&v_i^\tau\neq v_j^\tau\end{array}\right.
    \label{eq: SR}
\end{equation}

These feature extraction approaches perform TLP through direct classification of the extracted features. Additionally, these approaches primarily focus on deriving one-step tasks based solely on the current snapshot ${G}_\tau$. These methods, while effective in certain scenarios, do not fully exploit the dynamics of temporal networks.

\subsubsection{Latent Variables}
\paragraph{Dynamic Graph Summarisation (DGS)} DGS is primarily a task-independent representation unit used to integrate historical graph snapshots into one comprehensive weighted snapshot. In the context of TLP, DGS plays a critical role in synthesizing historical network dynamics. 
DGS collapses successive historical snapshots $G^{\tau}_{\tau - \Delta^{\prime}}$ into a comprehensive weighted snapshot $\bar G^{\tau}_{\tau - \Delta^{\prime}}$ using some kernel functions to aggregate while ensuring that essential properties of the dynamic topology are preserved. The DGS process is given by: 
\begin{align}
    \bar G^{\tau}_{\tau - \Delta^{\prime}} &\equiv \alpha_{\tau-\Delta^{\prime}} G_{\tau-\Delta^{\prime}} + \alpha_{\tau-\Delta^{\prime}+1} G_{\tau-\Delta^{\prime}+1} + \cdots + \alpha_{\tau} G_{\tau} 
    = \sum_{t=\tau-\Delta^{\prime}}^\tau \mathrm{K}(G_t;\tau,\theta)
    \label{eq: DGS}
\end{align}

The work in~\cite{sharan_temporal-relational_2008,9382981} introduced DGS as a representation unit and compared it to traditional classification models, referred to as inference unit. They built the model based on the successive adjacency matrix $\mathbf{A}$.

The primary distinction among various DGS methods lies in their kernel functions, such as the exponential and linear functions. \citet{hill_building_2006} demonstrated DGS methods' applicability to TLP through direct inference.

\paragraph{Matrix Factorisation (MF)} MF, also known as matrix decomposition, decomposes historical adjacency matrices $\mathbf A^\tau_{\tau - \Delta^{\prime}}$ or their transformations to extract latent features of temporal networks. These features can then be combined with other inference units or make direct inferences of future snapshots.

Several matrix factorisation techniques have been applied to TLP with considerable success. These include Eigenvalue Decomposition (ED), Singular Value Decomposition (SVD), Tensor Factorisation (TF), and Non-negative Matrix Factorisation (NMF), all of which are frequently used as task-dependent dynamic network embedding methods for TLP.

\textbf{Eigenvalue Decomposition (ED)} ED~\cite{wu_tracking_2018} utilises spectral graph theory to mine latent variables for tracking and predicting temporal networks. Spectral analysis involves the ED of the Laplacian matrix for each snapshot. In graph theory, each eigenvalue can be seen as a specific frequency revealing particular aspects of the graph's structure, such as its connectivity, robustness, and community structure. For a given timestamp $t$, the ED can be written as:
\begin{equation}
    \mathbf{A}_{t}=\mathbf{U}_t\boldsymbol{\Lambda}_{t}\mathbf{U}_t^{\top}
\end{equation}
where $\mathbf{A}_t$ is the adjacency matrix of a network, $\mathbf{U}_t$ is an orthogonal matrix, and $\boldsymbol{\Lambda}_t = \mathrm{diag}(\lambda_1^t, \lambda_2^t, \dots, \lambda_{N}^t)$ is the eigenvalue diagonal matrix at the timestamp $t$. Here, $N$ denotes the number of nodes in the network, which also corresponds to the dimension of the adjacency matrix $\mathbf{A}_t$.
\citet{kunegis_network_2010} studied spectral evolution based on ED and took advantage of two successive timestamps to solve TLP. 

\textbf{Singular Value Decomposition (SVD)} SVD is a generalised Eigenvalue Decomposition (ED) that extends to non-square matrices. SVD is commonly integrated with spectral analysis or dynamic graph summarisation to generate embeddings for TLP. For a collapsing adjacency matrix, $\bar{\mathbf{A}}_{\tau - \Delta^{\prime}}^\tau$ of dynamic graph summarisation, its SVD form is:
\begin{equation}
  \bar{\mathbf{A}}_{\tau - \Delta^{\prime}}^\tau = \mathbf{U}\boldsymbol{\Sigma}\mathbf{V}^{\top}
\label{eq: SVD}  
\end{equation}
where $\mathbf{U}$ and $\mathbf{V}$ are orthogonal matrices, and $\boldsymbol{\Sigma}$ is a diagonal matrix containing the singular values. Typically, $\mathbf U$ is the target embedding of the historical graph, while $\mathbf V$ and $\boldsymbol{\Sigma}$ are supplementary matrices.

\textbf{Truncated SVD (TSVD)}  TSVD is a variant of SVD that provides a low-rank approximation of the input matrix. Applying a $K$-rank approximation, the TSVD of $\bar{\mathbf{A}}_{\tau - \Delta^{\prime}}^\tau$ can be given by:
\begin{equation}
    \bar{\mathbf{A}}_{\tau - \Delta^{\prime}}^\tau \approx \mathbf{U}_K \boldsymbol{\Sigma}_K \mathbf{V}_K^{\top}
\label{eq: TSVD}
\end{equation}
where $\mathbf{U}_K$ and $\mathbf{V}_K$ are matrices formed from the first $K$ columns of $\mathbf{U}$ and $\mathbf{V}$, respectively, and $\boldsymbol{\Sigma}_K$ is the principal $K \times K$ submatrix of $\boldsymbol{\Sigma}$. The work in~\cite{acar_link_2009,dunlavy_temporal_2011} leveraged the Katz Index~\cite{katz_new_1953} based on the collapsing adjacency matrix to generate a similarity score matrix for TLP.

LIST~\cite{yu_link_2017} and TMF~\cite{yu_temporally_2017} both employ SVD to analyse network dynamics. LIST integrates spatial and temporal data for both one-step and multi-step tasks, using gradient descent to optimise its inference unit. This methodology is advantageous for temporal networks with diverse structural properties but may introduce significant computational overhead during the optimisation process. TMF is less computationally demanding, making it more efficient but potentially less effective in networks where spatial relationships are key and only suitable for multi-step tasks.

\textbf{Tensor Factorisation (TF)} TF is derived from SVD. The CanDecomp/Parafact (CP) decomposition is one of the most important TF methods for TLP~\cite{faber_recent_2003}. The work in~\cite{acar_link_2009,dunlavy_temporal_2011} examined homogeneous and bipartite graphs, such as user-item networks. These graphs have nodes and edges of the same type, divided into two disjoint sets of size $m$ and $n$ (where $m+n=N$), such that their sum equals the total number of nodes, $N$. They defined the relationship matrix $\mathbf{R}$ of size $m \times n$, where each element $(\mathbf{R})_{i,j}$ represents the relationship between $v_i$ and $v_j$:
\begin{equation}
(\mathbf{R})_{i,j}=\begin{cases}1&\text{if $v_i$ links to $v_j$}\\ 0&\text{otherwise}\end{cases}
\label{eq: Relationship matrix}
\end{equation}
For temporal networks, this relationship matrix becomes $\mathbf{R}_t$ at timestamp $t$, and extends to a tensor $\mathcal{R}^{m \times n \times |\mathcal{T}|}$ across all snapshots. 

TBNS~\cite{yang_link_2012} utilises tensor factorisation as the representation unit, capturing time series trends in network data. TBNS processes network snapshots into a two-dimensional matrix using exponential smoothing, enabling effective node similarity calculations for link prediction. However, TBNS generates sparse matrices, which may result in storage inefficiencies.

\textbf{Non-negative Matrix Factorisation (NMF)} \citet{gao_temporal_2011} introduced NMF for TLP. \citet{huang_non-negative_2012} showed that NMF decomposes a non-negative matrix such as an adjacency matrix $\mathbf{A}_t$ into two non-negative matrices $\mathbf{U}_t \in \mathbb R^{N_t \times d}_+$ and $\mathbf{V}_t \in \mathbb R^{N_t \times d}_+$. Here $d$ is the dimension of the latent space $(d < N_t)$. Their product approximates the original matrix: 
\begin{equation}
    \mathbf{A}_t\approx \mathbf U_t \mathbf V^{\top}_t
    \label{eq: NMF}
\end{equation}
where $\mathbf{U}_t$ is the target embedding matrix. This method is also known as classic bi-factor NMF~\cite{huang_non-negative_2012}. It needs to optimise the following Frobenius norm $\|\cdot\|_F$ of each snapshot:
\begin{equation}
\operatorname*{arg\,min}\limits_{\mathbf U_t \geq 0, \mathbf V_t \geq 0} J_\text{BNMF} \equiv \| \mathbf{A}_t - \mathbf U_t \mathbf V^{\top}_t\|_F^2
\label{eq: Bi-NMF}
\end{equation}

CRJMF~\cite{gao_temporal_2011} and AM-NMF~\cite{lei_adaptive_2018} comprehensively integrate temporal snapshots with node attributes. CRJMF utilises a tri-factorisation method, which is thorough but computationally intensive. AM-NMF combines matrix factorisation from consecutive snapshots, balancing historical and recent data through a decay factor for TLP.

SNMF-FC~\cite{ma_nonnegative_2017} and DeepEye~\cite{ahmed_deepeye_2018} prioritise computational efficiency by simplifying the modelling process. SNMF-FC focuses on temporal data through a decayed summation of feature matrices, potentially overlooking spatial details. DeepEye stabilises network dynamics across snapshots with consensus matrices, efficient in networks with gradual changes but possibly missing abrupt shifts.

GrNMF~\cite{ma_graph_2018} extends SNMF-FC by incorporating graph regularisation to better maintain spatial relationships among nodes. This approach enhances the model’s ability to capture both spatial and temporal information albeit with increased computational demands.

In summary, CRJMF and GrNMF are suitable for detailed spatial-temporal analysis but require significant computational resources. SNMF-FC and DeepEye provide efficient alternatives for less dynamic networks, while AM-NMF's adaptable framework makes it versatile for a variety of network analysis scenarios, balancing detail with computational efficiency.

\paragraph{Random Walk} In a static network, a classic random walk is a sequence of nodes where each vertex is randomly chosen from the neighbours of the current node. The random walk starts at an initial vertex and continues by selecting a neighbouring vertex with a certain transition probability and length $L$ to form a walk $W=(v^{t(1)}_a,v^{t(2)}_b,\dots,v^{t(L)}_c)$ such that the $r$-th node $v^{t(r)}_i \in V_t$ and edge $(v^{t(r)}_i,v^{t(r+1)}_j) \in E_t$.

There are static and dynamic approaches for the random walk in discrete-time dynamic graphs for TLP. Static random walk methods model each snapshot independently to generate embeddings, which are then used by the representation units for TLP. Well-known algorithms that apply this approach include DeepWalk~\cite{perozzi_deepwalk_2014} and Node2Vec~\cite{grover_node2vec_2016,de_winter_combining_2018}. In contrast, dynamic random walk methods directly model the dynamics of the network such as DynNode2Vec~\cite{mahdavi_dynnode2vec_2018} and SGNE~\cite{du_dynamic_2018}.

\textbf{Static Methods}
DeepWalk~\cite{perozzi_deepwalk_2014} and Node2Vec~\cite{grover_node2vec_2016} are influential graph analysis methods that utilise random walks to learn node embeddings, drawing inspiration from Skip-gram model~\cite{mikolov_efficient_2013}. DeepWalk generates node sequences through random walks and employs the Skip-gram model to capture higher-order proximity between nodes, effectively preserving local connectivity patterns. In contrast, Node2Vec extends DeepWalk's methodology by introducing biased random walks that balance breadth-first and depth-first search~\cite{moore_shortest_1959, tarjan_depth-first_1972}, enabling the model to capture both local and global topological information. This makes Node2Vec particularly adept at exploring complex network structures.

\citet{de_winter_combining_2018} first applied Node2Vec to temporal networks by using Node2Vec on each snapshot independently to generate an embedding sequence for each node. Then, they utilised traditional classifiers as inference units to make TLP, including Logistic Regression~\cite{cox_regression_1958}, Random Forests~\cite{breiman_random_2001}, and Gradient Boosted Regression Trees~\cite{shi_gradient_2018}. However, these static methods have issues with time-consuming computations and consistency.

In summary, Node2Vec offers a more flexible approach than DeepWalk by allowing for adjustable exploration of a node's neighbourhood. However, both methods face scalability issues and computational challenges when applied to temporal networks, necessitating considerations of efficiency and dynamic consistency in their application.

\textbf{Dynamic Methods}~\citet{du_dynamic_2018} proposed SGNE, extending the Skip-gram algorithm based on DynNode2Vec~\cite{mahdavi_dynnode2vec_2018}. While both methods adapt Skip-gram to capture temporal dynamics in discrete-time representation, SGNE introduces two key innovations: a decomposable loss function for learning representations of new nodes, and a selection mechanism that identifies the most influential nodes affected by network evolution. The focus on influential nodes potentially enables SGNE to achieve higher accuracy in TLP, particularly in networks where specific nodes drive structural changes.

\paragraph{Autoencoder} Autoencoder usually consists of encoders and decoders. The encoder transforms inputs into the latent representation space, whereas the decoder maps these embeddings back to obtain reconstructed inputs. Embeddings can be learned by minimising the reconstruction loss function. Nowadays, autoencoders are widely implemented by various kinds of neural networks.

DynGEM~\cite{goyal_dyngem_2018}, SDNE~\cite{wang_structural_2016}, and Dyn-VGAE~\cite{mahdavi_dynamic_2020} are three distinct approaches to dynamic node embeddings, each utilising variations of autoencoders to adapt to network evolution. DynGEM leverages a deep autoencoder, incrementally updating embeddings from one snapshot to the next and dynamically adjusting its architecture to accommodate growing network sizes. This flexibility in handling network expansion is advantageous but may lead to increased complexity in the neural network configuration. SDNE~\cite{wang_structural_2016}, a semi-supervised model, reconstructs the adjacency matrix to maintain first-order and second-order proximities, which helps preserve local and global topological features. This dual focus allows SDNE to maintain high fidelity in representing node relationships but at the cost of potentially higher computational demands due to the semi-supervised nature of the loss function. Dyn-VGAE~\cite{mahdavi_dynamic_2020} advances this by 
integrating Variational Graph Autoencoders (VGAE)~\cite{kipf_variational_2016} with Kullback-Leibler divergence~\cite{phoenix_elements_1992}, optimising embeddings for current snapshots while ensuring temporal consistency. This approach provides a robust framework for temporal coherence but requires careful tuning of hyper-parameters to balance immediate embedding accuracy with longitudinal consistency. Overall, each model presents a unique strategy to handle dynamic networks, with varying degrees of complexity and focus on either architectural flexibility, topological fidelity, or temporal stability.

\paragraph{Restricted Boltzmann Machine-based Approaches}
The Restricted Boltzmann Machine (RBM) is a deep learning structure originating from the Markov random field. It has two layers: a visible layer $\mathbf v \in \mathbb{R}^{N \times 1}$ and a hidden layer $\mathbf h \in \mathbb{R}^{d \times 1}$. The elements in $\mathbf v$ and $\mathbf h$ are stochastic binary variables representing observable data and latent representations~\cite{6796673}. RBM defines a joint distribution over $\mathbf v$ and $\mathbf h$ through the energy function:
\begin{equation}
\Pr(\mathbf{v},\mathbf{h})=\frac{\exp\{\mathbf{v}^\top \mathbf{W}\mathbf{h}+\mathbf a^\top\mathbf{v}+\mathbf b^\top\mathbf{h}\}}{\sum_{\mathbf{v},\mathbf{h}}\exp\{\mathbf{v}^\top \mathbf{W}\mathbf{h}+\mathbf a^\top\mathbf{v}+\mathbf b^\top\mathbf{h}\}}
\label{eq: RBM}
\end{equation}
where $\mathbf W \in \mathbb{R}^{N \times d}$ is the weight matrix; the bias vectors $\mathbf a \in \mathbb{R}^{N \times 1}$ and $\mathbf b \in \mathbb{R}^{d \times 1}$ correspond to $\mathbf{v}$ and $\mathbf h$ respectively. The loss function $J_{\text{RBM}}$ aims to minimise the negative log-likelihood:
\begin{equation}
    \min_{\mathbf{W}} J_\text{RBM}(\mathbf{W};\mathbf{v,h})\equiv-\ln \left(\sum_\mathbf{h}\Pr(\mathbf{v},\mathbf{h}) \right)
    \label{eq: RBM loss}
\end{equation}
ctRBM~\cite{li_deep_nodate} can capture complex non-linear variations using an exponentially large state space. ctRBM consists of two separate layers of visible units, the historical layer $\tilde{\mathbf{v}}$ and the predictive layer $\mathbf{v}$, both fully connected to hidden units $\mathbf h$. These layers receive inputs from the historical topology $G^{\tau}_{\tau-\Delta^\prime+1}$ and the prediction result $\hat{G}_{\tau+1}$ (or the training ground truth), respectively.  ctRBM makes a prediction based on the current time window $(\tau-\Delta^\prime,\tau]$ and local neighbour. For each node $v_i$, define $\tilde{\mathbf{v}} = ((\mathbf A_{\tau-\Delta^\prime+1})_{i,:}, \dots,(\mathbf A_{\tau})_{i,:})^\top \in \mathbb{R}^{N \Delta^\prime \times 1}$ and $\mathbf{v} = (\mathbf A_{\tau+1})_{i,:}^\top \in \mathbb{R}^{N \times 1}$.
ctRBM uses direct inference to predict the future adjacency matrix $\tilde{\mathbf A}_{\tau+1}$:
\begin{equation}
    (\hat{\mathbf A}_{\tau+1})^\top_{i,:} =f_\text{TT}(G^{\tau}_{\tau-\Delta^\prime}) \equiv \Pr(\mathbf{v}\mid \mathbf{h};\delta)=\sigma(\mathbf{Wh}+\mathbf{a})
\end{equation}

\paragraph{Graph Neural Networks-based Discrete-time Approaches}
Graph neural networks (GNNs) are a type of neural networks that operate on graph-structured data. As one of the most popular models in recent years, GNNs possess powerful abilities in graph representation learning~\cite{khoshraftar_survey_2022}. Most GNNs are based on the message-passing mechanism. This powerful mechanism consists of memorizing and aggregating the messages of nodes in the network. Therefore, GNNs can capture the dynamics in the temporal networks and update the embeddings of nodes and edges accordingly. Apart from their representation ability, the flexible architectures of GNNs help them integrate with various other dynamic network embedding methods. Thus, GNNs can handle both discrete-time and continuous-time dynamic graphs. Same as random walks, GNNs could also be divided into static and dynamic methods for discrete-time dynamic graphs. 

\textbf{Static Methods}
Several GNNs have been proposed and achieved impressive results on static networks such as Graph Convolutional Network (GCN)~\cite{kipf_semi-supervised_2017}, LINE~\cite{tang_line_2015} and GraphSAGE~\cite{hamilton_inductive_2017}.  They all generate node embeddings. GCN uses convolutions to aggregate information from node neighbours. LINE preserves first-order and second-order proximities between nodes to learn embeddings. GraphSAGE is based on GCN and uses a sampling technique to handle large graphs. 

Spatial-temporal GNNs (STGNNs) are a type of GNN that extends traditional static GNNs to handle temporal networks with spatial information~\cite{bui_spatial-temporal_2022,sahili2023spatiotemporalgraphneuralnetworks}. STGNNs constitute a distinct class of GNNs. These networks model the dynamics by accounting for dependencies between connected nodes. STGNNs are widely applied in traffic forecasting~\cite{zhang_spatio-temporal_2020,zhang_traffic_2021,ali_test-gcn_2021}, and epidemic prediction~\cite{wang_causalgnn_2022}.

\citet{wu_graph_2019}~proposed Graph WaveNet by extending the Convolutional Neural Network (CNN)~\cite{lecun_convolutional_1995,oshea_introduction_2015} to an STGNN. The Graph WaveNet consists of a spatial convolution layer and a temporal convolution layer. The spatial convolution layer combines a diffusion convolution layer~\cite{li_diffusion_2017} with a self-adaptive adjacency matrix, while the temporal convolution layer adopts a gated version of a dilated causal convolution network \cite{yu_multi-scale_2015}. 

\textbf{Dynamic Methods} 
Dynamic GNNs are directly designed to handle discrete-time dynamic graphs. There are also several other dynamic GNNs based on the Graph Convolutional Network (GCN). For example,  Stochastic Gradient Descent (SGD)~\cite{ruder_overview_2016} combines GCN with PageRank similarity~\cite{gasteiger_predict_2018,bojchevski_scaling_2020,fu_sdg_2021}.

DyGCN~\cite{cui_dygcn_2022} is a typical task-independent dynamic network embedding method based on GCN to address the challenges of temporal networks. The key of DyGCN is to generalise the embedding propagation scheme of GCN to a dynamic setting in an efficient manner to update the node embedding matrix $\mathbf{Z_{\tau}}$. DyGCN assumes the feature matrix $\mathbf{X}$ is fixed and the node embeddings are updated according to the change of aggregated message. 

GCN-MA~\cite{mei_dynamic_2024} uses a GCN as the representation unit. GCN-MA includes the novel Node Representation Node Aggregation Effect (NRNAE) algorithm which is a combination of GCN, RNN, multi-head attention mechanisms, enhancing node representation through node degree, clustering coefficient, and neighbour relationships. Thus, GCN-MA has an improved ability to capture global and local temporal patterns.

\subsubsection{Summary}
Discrete-time dynamic graph approaches transform temporal networks into sequences of static snapshots, enabling the extraction of meaningful representations for TLP. These representations can either be integrated with specialised inference units or directly applied to prediction tasks.
The methods in this category include dynamic graph summarisation, matrix factorisation, and random walk techniques, each capturing distinct aspects of graph structure and temporal evolution. Neural approaches such as autoencoders provide efficient data compression, whilst Restricted Boltzmann Machines and Graph Neural Networks (GNNs) offer powerful modelling of structural and temporal patterns.

\subsection{Continuous-time Dynamic Graphs Methods}
\subsubsection{Latent Variables}
\paragraph{Graph Neural Networks-based Continuous-time Approaches}
\textbf{Traditional GNN}
TGAT~\cite{xu2020inductiverepresentationlearningtemporal} extends the dynamic network embedding ability of GAT~\cite{velickovic_graph_2018} via Function Time Encoding. GAT is used for static settings and does not consider the temporal dynamics between neighbours. To process continuous-time dynamic graphs, the time features used in TGAT are derived from concepts based on Bochner's Theorem to map time to $\mathbb{R}^d$. Then, $\Phi(t)$ is concatenated with node embedding for GAT to solve TLP. However, TGAT cannot maintain the historical state of the nodes.

TGN~\cite{rossi_temporal_2020} can memorise the history information to generate node embeddings for continuous-time dynamic graphs. The model comprises several modules, including Memory, Message Function, Message Aggregator, Memory Updater, and Embedding. The Memory module stores a vector for each node representing the node's history in a compressed format at a certain timestamp. Other modules implement the message-passing mechanism to generate temporal node embedding. Therefore, TGN performs better than the TGAT.

TGGDN~\cite{huang_temporal_2023} employs a group affinity matrix to model both local and long-distance interactions within networks, incorporating a transformer architecture for temporal data processing and enhanced interpretability. CoDyG~\cite{chen_correlation-enhanced_2024} introduces a co-attention mechanism alongside a temporal difference encoding strategy to effectively capture evolving correlations between node pairs over time.

These approaches represent distinctive methodologies for modelling temporal graph dynamics, with TGGDN focusing on group-level interactions and broader network patterns, whilst CoDyG emphasises the fine-grained temporal evolution of pairwise node relationships.

\textbf{Temporal Point Process with GNN}
The temporal point process is a probabilistic generative model for continuous-time event sequences, which involves a stochastic process whose realization consists of a list of discrete events localised in time~\cite{shchur_neural_2021}. Assuming an event happens within a tiny period $[t, t + dt)$, the temporal point process represents the conditional probability of this event given historical events as $\lambda(t)dt$. The Hawkes process~\cite{hawkes_spectra_1971} is a widely-used temporal point process method for TLP. The Hawkes process is described by the equation \cite{yuan_multivariate_2019}:
\begin{equation}
    \lambda(t) = \mu(t) + \int_{-\infty}^{t} \kappa(t-s)dE(s)
    \label{eq: Hawkes}
\end{equation}
where the conditional intensity function $\lambda(t)$ represents the instantaneous rate of event occurrence at timestamp $t$; $\mu(t)$ represents the base intensity, which indicates the rate at which spontaneous events occur at timestamp $t$; $\kappa(t-s)$  is the kernel, modelling the time decay of past events' influence on the current intensity; $E(t)$ signifies the number of events up to timestamp $t$. This process incorporates the self-exciting and the past influence mechanisms of events, allowing it to effectively capture the complex dependencies and dynamics in continuous-time event sequences~\cite{saha_modeling_2021, morariu-patrichi_state-dependent_2022}.

\citet{zuo_embedding_2018} introduced HTNE, which captures network evolution by modelling how past interactions influence future connections through an attention mechanism that weighs historical interactions based on temporal proximity. 

Building upon this foundation, M\textsuperscript{2}DNE~\cite{lu_temporal_2019} incorporates both micro-dynamics (individual interactions) and macro-dynamics (subgraph changes) through a dual-attention mechanism that enhances TLP by balancing direct interactions with overarching network evolution patterns.

MTNE~\cite{huang_motif-preserving_2020} addresses limitations in previous approaches by modelling network dynamics through triad motif evolution and the Hawkes process. This method captures mesoscopic structural patterns neglected by HTNE and M\textsuperscript{2}DNE, which primarily focus on direct neighbour interactions and broad network changes, respectively.

TREND~\cite{wen_trend_2022} combines the Hawkes process with Temporal Graph Networks to model both event dynamics and node dynamics simultaneously. It characterises each edge formation as an event with properties determined by the participating nodes at specific timestamps.

These approaches contribute different methodological perspectives to continuous-time dynamic graph embedding: HTNE focuses on temporal relevance through attention mechanisms, M\textsuperscript{2}DNE addresses micro and macro network dynamics, MTNE captures mesoscopic structural patterns via motif evolution, and TREND combines event and node dynamics. Each method offers distinct techniques for modelling temporal network relationships, providing various analytical frameworks for understanding complex network behaviours.

\paragraph{Temporal Walk}

A temporal walk extends the concept of random walks~\cite{perozzi_deepwalk_2014, grover_node2vec_2016} to continuous-time dynamic graphs. Formally, a temporal walk is defined as a sequence of nodes $W_T = (v^{(1)}_a, v^{(2)}_b, \dots, v^{(L)}_c)$ with length $L$ during time domain $T$, where the $r$-th node $v^{(r)} \in V_T$, the $r$-th edge $((v^{(r)},v^{(r+1)}),t^{(r)})\in E_{T}$, and each timestamp $t^{(r)} \in T$. Crucially, timestamps must follow a temporal ordering (either ascending or descending) to properly capture network dynamics.

Unlike standard random walks, temporal walks feature irregular time intervals between steps whilst respecting the temporal ordering of connections, making them particularly suitable for continuous-time dynamic graphs.

CTDNE~\cite{nguyen_continuous-time_2018} leverages temporal walks for dynamic network embedding. The method first generates these walks using edge selection based on uniform, exponential, or linear distributions, then applies the Skip-Gram model~\cite{mikolov_distributed_2013} to learn the resulting embeddings. As a task-independent dynamic network embedding approach, CTDNE produces representations applicable across various temporal network tasks.

\paragraph{Neighbour Sequence} 

A Neighbour Sequence is an ordered collection of a node’s interactions with its one-hop neighbours over time. Each interaction in the sequence typically includes information such as the neighbour's identity and the time of interaction. By restricting attention to one-hop neighbours, models can focus on the most relevant local context whilst significantly reducing computational and storage requirements compared to methods that capture larger neighbourhoods or more intricate structures.

Recent works demonstrate the growing adoption of this strategy. For instance, GraphMixer~\cite{cong_we_2023} utilises a multi-layer perceptrons-based link encoder and a mean-pooling node encoder on the one-hop neighbour sequence, achieving outstanding link prediction performance with simpler architectures. DyGFormer~\cite{yu_towards_2023} employs a neighbour co-occurrence encoding scheme on neighbour sequences alongside a   Transformer based on attention mechanism~\cite{vaswani_attention_2017}, thereby efficiently capturing long-term dependencies. Building upon neighbour sequence, FreeDyG~\cite{tian_freedyg_2024} integrates frequency encoding to further exploit the node interactions, reinforcing the effectiveness of focusing on a node’s immediate neighbourhood whilst unveiling temporal frequency patterns.

\subsubsection{Summary}
GNN-based methods offer computational efficiency and effective retention of historical information, whilst temporal walk approaches excel at capturing relative positional relationships between nodes. Consequently, both methodologies can deliver outstanding performance.

\section{Review of Inference Units for TLP}\label{sec: literature2}
\subsection{Transductive Inference Methods}
\subsubsection{Direct Inference}

Direct inference is a widely used technique to solve TLP. This approach is characterised by its simplicity and directness, allowing the embeddings or features generated through dynamic network embedding methods to be utilised immediately without the need for further training in other inference units. Commonly implemented techniques in this context include matrix factorisation, dynamic graph summarisation, GTRBM~\cite{li_restricted_2018}, and ctRBM~\cite{li_deep_nodate}. These techniques are valued for their straightforward application in one-step tasks, making them one of the most popular choices in the field.

\subsubsection{Classification}

TLP can be framed as a binary classification problem, where the aim is to predict the presence or absence of a link between two nodes in the future. Almost all the methods that fall into this category rely on supervised learning. These classifiers usually utilise features or embeddings obtained from representation units for TLP. Various classifiers have been employed for link prediction including Logistic Regression~\cite{cox_regression_1958}, Support Vector Machine (SVM)~\cite{bliss_evolutionary_2014}, and Decision Trees (DT)~\cite{wang_human_2011}. 

TVRC~\cite{sharan_temporal-relational_2008} combines dynamic graph summarisation as the representation unit with Weighted Relational Bayes Classifiers \cite{neville_simple_2003} as the inference unit. TVRC takes the inputs from dynamic graph summarisation into Weighted Relational Bayes Classifiers (RBC)~\cite{neville_simple_2003} or Weighted Relational Probability Trees (RPT) \cite{neville_learning_2003} for TLP. RBC extends naive Bayes classifiers and RPT extends standard probability estimation trees to a relational setting. Both RBC and RPT incorporate weights from the collapsed snapshots from dynamic graph summarisation to better model the temporal networks.

GTRBM~\cite{li_restricted_2018} uses RBM as the representation unit and Gradient Boosting Decision Tree (GBDT)~\cite{friedman_greedy_2001} as the inference unit to form the model. GTRBM used the same RBM as ctRBM but the embeddings are input into the GBDT which has better inference ability than Bayes Classifier. As a result, GTRBM provides better performance than ctRBM which relies on direct inference for TLP.

\subsubsection{Time Series}
Temporal networks can be conceptualised as time-series data, allowing TLP problems to be approached through established time-series methodologies. These approaches vary based on their applicability to discrete-time or continuous-time dynamic graphs.

\paragraph{For discrete-time dynamic graphs} 
Classical time series techniques including Auto-Regressive (AR)~\cite{kalman_new_1960}, Auto-Regressive Moving Average (ARMA), Auto-Regressive Integrated Moving Average (ARIMA)~\cite{cholette_prior_1982,brockwell_introduction_2016}, and Vector Auto-Regressive (VAR)~\cite{lutkepohl_new_2005}, form the backbone of many TLP models when combined with various representation units.

Feature extraction methods incorporate similarity measures such as Common Neighbours, Adamic-Adar and Jaccard Coefficient~\cite{Newman2001ClusteringAP,adamic_friends_2003,Goodall1978} with time series analysis. For example, TSalton~\cite{zhang_temporal_2020} integrates ARIMA as its inference unit to predict node centrality, thereby enhancing TLP adaptability dynamically.

Matrix factorisation approaches such as SimRank (SR)~\cite{fang_graph_2010} and TRMF~\cite{yu_temporal_2016} apply time series principles to dynamic matrix representations. SR combines spectral analysis with ARMA to predict future eigenvectors for graph reconstruction, employing $K$-rank approximation as Equation~\eqref{eq: TSVD} to address computational challenges in large networks. Similarly, TRMF pairs Non-negative Matrix Factorisation with AR to predict temporal embeddings, effectively adapting matrix factorisation to dynamic contexts.

These methods offer different strengths: feature extraction approaches like TSalton directly incorporate temporal dynamics into node similarity metrics, whilst matrix factorisation methods capture the structural information. Computational requirements vary significantly, with spectral methods requiring dimensional simplifications and NMF methods offering more straightforward implementation but potentially less structural detail.

\paragraph{For continuous-time dynamic graphs} 
Continuous-time methods expand the traditional Discrete-time approaches such as AR and ARMA to Continuous-time AR (CAR)~\cite{harvey_continuous_1988} and Continuous-time ARMA (CARMA)~\cite{brockwell_use_1996}. Moreover, \cite{xu_real-time_2017} combined ARIMA and Kalman filtering as the ARIMA-Kalman model for continuous-time dynamic graphs.  The ARIMA model is employed to initialise the Kalman measurement and state equations. The model combines the linear pattern-capturing strengths of ARIMA and the adaptive noise reduction ability of Kalman filtering. Therefore, ARIMA–Kalman improves the flexibility and forecasting accuracy of TLP.

\subsubsection{Summary}
Transductive inference methods employ graph-based representations for specific tasks such as classification and time series analysis. While these approaches effectively utilise structural and temporal information for predictive modelling, they struggle to generalise beyond observed samples, highlighting limitations in their predictive capabilities. Moreover, most of these inference units are only compatible with representation units in discrete-time dynamic graphs.

\subsection{Inductive Inference Methods}

\subsubsection{Recurrent Neural Networks-based Approaches}
Recurrent Neural Networks (RNNs) refer to a class of neural networks where connections between nodes can form cycles, enabling output from some nodes to influence subsequent input to the same nodes. This allows RNNs to model sequential data and process the dynamics of temporal networks. RNNs have already succeeded in many other tasks such as speech recognition or language translation.

However, traditional RNNs suffer from gradient descent and gradient explosion problems. Consequently, RNNs may be unable to capture long-range data dynamics dependencies. To overcome these issues, Long Short-Term Memory (LSTM) \cite{gers_learning_2000} and Gated Recurrent Unit (GRU)~\cite{chung2014empiricalevaluationgatedrecurrent} networks were developed based on RNNs to handle long-term learning issues. RNNs could also be divided into discrete-time and continuous-time methods.

\paragraph{For discrete-time dynamic graphs} 
DDNE~\cite{li_deep_2018} uses adjacency matrices to create time-specific embeddings. DDNE employs two Gated Recurrent Units (GRUs) to analyse these embeddings, capturing the network's evolution over time. This method enables the generation of advanced embeddings and dynamic inferences using Multi-Layer Perceptions (MLP). The dual GRU -- one processing data from past to present, the other in reverse -- allows DDNE to understand temporal dependencies within networks comprehensively. This facilitates accurate TLPs, showcasing DDNE's capability to harness network dynamics effectively. 

DynGraph2Vec~\cite{goyal_dyngraph2vec_2020}  uses an autoencoder as the representation unit and RNN as the inference unit. The method processes adjacency matrices from previous time steps, employing a Multi-Layer Perceptron (MLP) as the encoder to transform these matrices into embeddings. This is followed by an LSTM-based decoder focused on TLP. DynGraph2Vec introduces variants such as DynGraph2VecAE, DynGraph2VecAERNN, and DynGraph2VecRNN to cater to different requirements and optimisation strategies.

The encoding process leverages MLP to convert the series of adjacency matrices over time into a sequence of node embeddings, effectively capturing the graph's evolving structure. The decoder, on the other hand, uses an LSTM layer fed by the generated embeddings to predict future adjacency matrices as TLP. This approach allows DynGraph2Vec to understand and anticipate the dynamic changes in graph structures, facilitating accurate embeddings for temporal networks.

EvolveGCN~\cite{pareja_evolvegcn_2020} utilises a Graph Convolutional Network (GCN) module as the representation unit to learn node embeddings in temporal networks. EvolveGCN employs a GRU to update the weights of the GCN for inference at each timestamp, allowing the model to capture the latent dynamic patterns of the temporal networks. For TLP, it employs a Multi-Layer Perceptron (MLP) that uses the final GCN-generated embeddings to predict future connections, which efficiently captures and predicts dynamics in temporal networks. 

GCN-GAN~\cite{lei_gcn-gan_2019} leverages GCN as the representation unit and LSTM as the inference unit under the architecture of Generative Adversarial Network (GAN)~\cite{mirza_conditional_2014,10.5555/3305381.3305404,goodfellow_generative_2020}. The GAN contains a generator and a discriminator. The generator contains the representation unit and the inference unit to make TLP, while the discriminator is an auxiliary structure to refine prediction results given by the generator. Similarly, NetworkGAN~\cite{yang_advanced_2020} improved the GCN-GAN model with matrix factorisation-based representation units.

\paragraph{For continuous-time dynamic graphs}
DyGNN~\cite{ma_streaming_2020} tracks the effects of new edges in temporal networks, blending GNN and LSTM to generate and update embeddings. This model captures both direct interactions and the influence on adjacent nodes through Continuous-time LSTM. A subsequent LSTM layer, equipped with attention and decay mechanisms, spreads the updates across the network, ensuring dynamic and precise embeddings. DyGNN excels in reflecting network changes and improving TLP. 

CAW-N is an anonymised variant of the temporal walk, capturing motif evolution in temporal networks. It operates on the premise that nodes with interacting motif structures over time have a higher likelihood of forming links. CAW-N employs temporal walks as representation units to track events between node motifs chronologically, then utilises RNNs for prediction.

\subsubsection{Attention-based Approaches}
Attention-based models refer to a set of methods that can mimic cognitive attention, allowing the models to focus on the important parts of the data while diminishing less relevant parts~\cite{soydaner_attention_2022}. These models, such as the global or local attention~\cite{xu_show_2015}, self/multi-head attention~\cite{vaswani_attention_2017}, Simple Neural Attentive Learner (SNAIL)~\cite{mishra2018simpleneuralattentivemetalearner} and Cardinality Preserved Attention (CPA)~\cite{wang_multi-hop_2020}, have gained popularity in many fields including Computer Vision and Natural Language Processing and have achieved state-of-the-art results in various tasks. One well-known model in graph learning is the Graph Attention Network (GAT)~\cite{velickovic_graph_2018}. GAT uses GNNs as the representation unit to process the graph structure and attention mechanism to improve the updating process of node embeddings.

\paragraph{For discrete-time dynamic graphs} 
DySAT~\cite{sankar_dysat_2020} employs a multi-layer GAT~\cite{vaswani_attention_2017} as its representation unit alongside a self-attention module for inference. The inductive nature of attention models enables DySAT to generalise to previously unseen nodes, contributing to the increasing popularity of attention-based approaches in temporal graph analysis.

Building on this foundation, ASTGCN~\cite{guo_attention_2019} and STGSN~\cite{min_stgsn_2021} further develop spatial-temporal modelling techniques. ASTGCN implements distinct attention modules to capture various temporal patterns through spatial-temporal mechanisms and convolution techniques. Similarly, STGSN utilises temporal attention mechanisms to model network evolution from both spatial and temporal perspectives, enhancing the interpretability of dynamic interactions.

NeiDyHNE~\cite{wei_neighbor-enhanced_2024} extends these concepts to heterogeneous networks by integrating neighbourhood interactions with node attributes. It combines a hierarchical structure attention module for analysing node features with a convolutional temporal attention module that captures evolutionary patterns. This integrated approach enables effective management of complex dynamics in heterogeneous networks, improving predictive performance for future connections.

\paragraph{For continuous-time dynamic graphs} 
Whilst many approaches employ continuous-time GNNs with attention mechanisms for transductive tasks, alternative methods can directly embed continuous-time dynamic graphs using attention models, such as those based on the deep learning Hawkes process described earlier by Equation~\eqref{eq: Hawkes}.

MAGNA~\cite{wang_multi-hop_2020} extends the attention mechanism from GAT~\cite{vaswani_attention_2017} by computing preliminary node embeddings using 1-hop attention matrices, then incorporates multi-hop neighbours through summed powers of these matrices. The model aggregates node features weighted by attention values and processes them through a feed-forward neural network to generate embeddings for TLP.

DyRep~\cite{trivedi_dyrep_2019} takes a different approach based on temporal point processes, modelling temporal networks through two evolving processes: the association process (capturing global network growth) and the communication process (representing local information propagation). DyRep employs the Hawkes process as its representation unit and applies attention mechanisms to compute edge embeddings for TLP.

\subsubsection{Summary}
The inference units for inductive tasks in TLP rely primarily on deep learning architectures, particularly RNN-based and attention-based models capable of generalising to unseen data. These architectures offer considerable flexibility, enabling them to either directly perform TLP after learning dynamic patterns or generate embeddings for downstream components. Their compatibility with both discrete-time and continuous-time dynamic graph representations demonstrates their effectiveness in capturing diverse temporal and structural patterns.

\section{Variations of the TLP Problem}\label{sec: variations}

This section explores extensions of TLP, from undirected homogeneous temporal networks to more complex variants, including directed, heterogeneous and hypergraph temporal networks, along with their applications. All network definitions are founded on discrete-time dynamic graph principles.

\paragraph{Link Prediction in Directed Temporal Networks}
A directed temporal network features links that evolve over time with specific directionality between nodes~\cite{badie-modiri_directed_2022,lv_graph_2022}. In a temporal network $\mathcal{G = (V, E, T, X)}$ where ${\mathcal{E}}\subseteq{\mathcal{V}}\times{\mathcal{V}}$, the edge $(v_i^t, v_j^t)$ indicates a connection from node $v_i^t$ to node $v_j^t$, distinct from edges in weighted homogeneous attribute temporal networks. These networks model phenomena including rumour or disease propagation via social networks \cite{daley_epidemics_1964,dietz_epidemics_1967}, ad hoc message passing~\cite{aguilar_igartua_special_2015} and public transport dynamics~\cite{nassir_utility-based_2016}. Temporal Knowledge Graphs represent a particularly significant application.

\paragraph{Temporal Knowledge Graph}
A Temporal Knowledge Graph is a special type of knowledge graph that incorporates time-varying information into the graph, representing the dynamics of entities and relationships in the knowledge graph~\cite{cai_temporal_2022}.
In Temporal Knowledge Graphs, methods for TLP need to consider both the knowledge and evolution of networks~\cite{wang_survey_2021,zhang_temporal_2022}. By incorporating time-aware representation learning models, Temporal Knowledge Graphs can reason missing temporal facts and relationships~\cite{zuo_learning_2022,cai_temporal_2022}. Various models have been proposed for TLP in Temporal Knowledge Graphs~\cite{gangemi_modeling_2018,jung_learning_2021} or combined with some Natural Language Processing techniques such as contrastive learning~\cite{liu_tlogic_2022} or temporal logic~\cite{liu_tlogic_2022}.

\paragraph{Link Prediction in Heterogeneous Temporal Networks}
A heterogeneous temporal network is a network whose nodes and edges are of different types and change over time~\cite{jaya_lakshmi_link_2017}.
Given a heterogeneous temporal network  $\mathcal{G} = (\mathcal{V}, \mathcal{E},\mathcal{T})$, $\mathcal{V} = \bigcup_{i=1}^{n} \mathcal{V}_i$ represents $n$ types of nodes, and $\mathcal{E} = \bigcup_{j=1}^{m} \mathcal{E}_j$ denotes $m$ types of edges. Examples of heterogeneous temporal networks include complex social networks~\cite{dong_link_2012} and recommendation systems~\cite{xie_komen_2022}.

Collective Link Prediction refers to a set of methods used for predicting the probability of new relationships forming between nodes in heterogeneous temporal networks~\cite{yang_predicting_2012,negi_link_2016}, and it is often based on attention models~\cite{10.1145/3308558.3313562,hutter_modeling_2021,jose_dynamic_2020}.

\paragraph{Link Prediction in Temporal Hypergraph Networks}
A temporal hypergraph network is a network whose edges can join any number of nodes and  changes over time~\cite{Lee2021THyMeTH}. 
Temporal hypergraph networks can represent and analyse complex real-world systems such as economic transactions or transportation~\cite{fischer_visual_2021}. Recently, they have drawn attention and were studied on the TLP problem achieving good performance based on GNNs and attention-based models~\cite{sun_multi-level_2021, https://doi.org/10.1049/itr2.12130}.

\section{Future Research Directions}\label{sec: future}
TLP represents a vital area of research within complex network science and graph representation learning. This section explores emerging challenges and unresolved questions in the field.

\paragraph{Building new models for TLP based on different representation units and inference units} In light of the novel framework introduced by this survey, which is characterised by its composite nature, there is a compelling argument for the development of new models for TLP that harnesses the potential of innovative combinations. This framework, by its ability to amalgamate diverse techniques and methodologies, opens up the possibilities for tackling future TLP challenges in novel and efficacious ways. The essence of this approach lies in its flexibility and adaptability, encouraging the exploration of uncharted territories within the domain of TLP. By integrating different representation units and inference units, this composite framework sets the stage for a  holistic and nuanced understanding of network dynamics. As such, not only does it broaden the horizon for TLP research but also serves as a beacon for future investigations, guiding them towards developing solutions that are both innovative and tailored to the complicated temporal network evolution.

\paragraph{Enhancing Explainability in TLP Models} A pivotal future direction emerges in the enhancement of model explainability. This initiative is critical for understanding what is viewed as the black box of TLP models, elevating their trustworthiness and deepening the comprehension of the mechanisms and dynamics in temporal networks. This direction does not merely aim to refine the predictive ability of TLP models but seeks to illuminate the underlying dynamics that characterise temporal networks. Therefore, this push for enhanced interpretability and explainability represents not just a trend towards innovation and integration within TLP frameworks but also emerges as a new direction aimed at developing models that balance explainability with capability. This direction may be one of the crucial foundations for future practical applications. 

\paragraph{Studying TLP on continuous-time dynamic graphs in complex networks.} TLP has been a classical task on weighted homogeneous attribute temporal networks for a long time. However, real-world temporal networks are often much more complex. They can be represented as directed, multi-layer, heterogeneous, or hypergraph networks. These complex networks pose challenges in TLP as their structure and evolution patterns are more intricate than the weighted homogeneous attribute temporal networks. Hence, this field remains an open challenge and requires novel techniques for such highly complex networks.

\paragraph{Further Exploration of the Graph Dynamics Mechanism.} Recent studies have examined mechanisms including Micro- and Macro-dynamics~\cite{lu_temporal_2019} and Motif structures~\cite{paranjape_motifs_2017,huang_motif-preserving_2020,yao_higher-order_2021,qiu_temporal_2023}. Nevertheless, a deeper understanding of complex network dynamics and how these mechanisms influence temporal network properties, such as varying speeds, remains elusive. Additionally, developing innovative approaches that leverage these mechanisms to more effectively capture evolution patterns in temporal networks is crucial.

\paragraph{General TLP Models in Large-scale Complex Networks.}  In recent years, the volume of data has been growing exponentially and the development of large models has been progressing rapidly~\cite{10.5555/3495724.3495883,kaplan2020scalinglawsneurallanguage}. For instance, social networks are becoming increasingly complex, encompassing various data types. Consequently, constructing a general and large-scale graph prediction model based on complex networks is becoming more and more promising. Such a model would have a strong ability in multi-task scenarios with domain knowledge, including recommendation systems, knowledge reasoning, epidemic spreading, etc. Future research in this area should focus on designing scalable and robust TLP models capable of processing large-scale and complex datasets and combining few-shot learning, transfer learning, and other advanced techniques to improve performance.

\section{Conclusion}\label{sec: conclusion}

This survey has introduced a novel taxonomy for Temporal Link Prediction (TLP) that distinguishes between representation units and inference units, providing a structured framework for analysing existing approaches. Through this lens, we have systematically reviewed and classified the literature, revealing the diverse methodological combinations employed across the field. The survey has also examined advanced TLP applications in directed, heterogeneous and hypergraph temporal networks, whilst identifying promising research directions including model explainability, complex network dynamics, and scalable solutions for large-scale networks.

The taxonomy framework presented here offers significant value to researchers by clarifying the fundamental components of TLP methods and highlighting unexplored combinations that could yield performance improvements. By emphasising the separation between representation and inference mechanisms, this work facilitates more targeted methodological innovations. Furthermore, our examination of the evolution from traditional feature-based approaches to advanced neural architectures provides a comprehensive roadmap for the field's development. We anticipate that this structured perspective will stimulate new research directions and accelerate progress towards more effective, interpretable and versatile TLP models capable of addressing the complex challenges inherent in temporal network analysis.

\backmatter

\section*{Declaration} 
The authors have no competing interests to declare that are relevant to the content of this article.


\end{document}